\pdfoutput=1
\documentclass[11pt]{article}

\usepackage[]{EMNLP2022}

\usepackage{times}
\usepackage{latexsym}

\usepackage[T1]{fontenc}
\usepackage[utf8]{inputenc}

\usepackage{microtype}

\usepackage{inconsolata}

\usepackage{booktabs}
\usepackage{fixltx2e}
\usepackage{amsmath,amsfonts,amssymb,mathtools}
\usepackage{graphicx}
\usepackage{multirow}
\usepackage{xcolor}
\usepackage{enumitem}
\usepackage{hyperref}
\usepackage{scalerel}
\usepackage{balance}
\usepackage{xspace}
\newcommand*{\scale}[2][4]{\scalebox{#1}{$#2$}}

\newcommand\blfootnote[1]{%
  \begingroup
  \renewcommand\thefootnote{}\footnote{#1}%
  \addtocounter{footnote}{-1}%
  \endgroup
}
\newcommand{\method}{\textsc{GraPe}\xspace}
\newcommand{\methodemoji}{\textsc{GraPe} \includegraphics[width=0.02\textwidth]{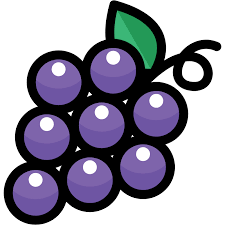}\xspace}
\newcommand{\methodemojititle}{\textsc{GraPe} \includegraphics[width=0.03\textwidth]{figures/grape.png}\xspace}

\title{\methodemojititle: Knowledge Graph Enhanced Passage Reader \\ for Open-domain Question Answering}

\author{{\bf Mingxuan Ju$^{1*}$, Wenhao Yu$^{1*}$, Tong Zhao$^2$, Chuxu Zhang$^3$, Yanfang Ye$^1$} \\ {$^{1}$University of Notre Dame, $^{2}$Snap Inc., $^{3}$Brandeis University} \\ $^{1}${$\{$\tt mju2,wyu1,yye7$\}$@nd.edu; \tt $^{2}$tzhao@snap.com; $^{3}$chuxuzhang@brandeis.edu}}

\begin{document}
\maketitle
\begin{abstract}
A common thread of open-domain question answering (QA) models employs a retriever-reader pipeline that first retrieves a handful of relevant passages from Wikipedia and then peruses the passages to produce an answer.
However, even state-of-the-art readers fail to capture the complex relationships between entities appearing in questions and retrieved passages, leading to answers that contradict the facts. In light of this, 
we propose a novel knowledge \underline{\textbf{Gra}}ph enhanced \underline{\textbf{p}}assag\underline{\textbf{e}} reader, namely \methodemoji, to improve the reader performance for open-domain QA.
Specifically, for each pair of question and retrieved passage, we first construct a localized bipartite graph, attributed to entity embeddings extracted from the intermediate layer of the reader model. Then, a graph neural network learns relational knowledge while fusing graph and contextual representations into the hidden states of the reader model.
Experiments on three open-domain QA benchmarks show \method can improve the state-of-the-art performance by up to 2.2 exact match score with a negligible overhead increase, with the same retriever and retrieved passages. Our code is publicly available at \url{https://github.com/jumxglhf/GRAPE}.
\end{abstract}

\section{Introduction}
\begin{figure}[t]
    \centering
    \includegraphics[width=0.49\textwidth]{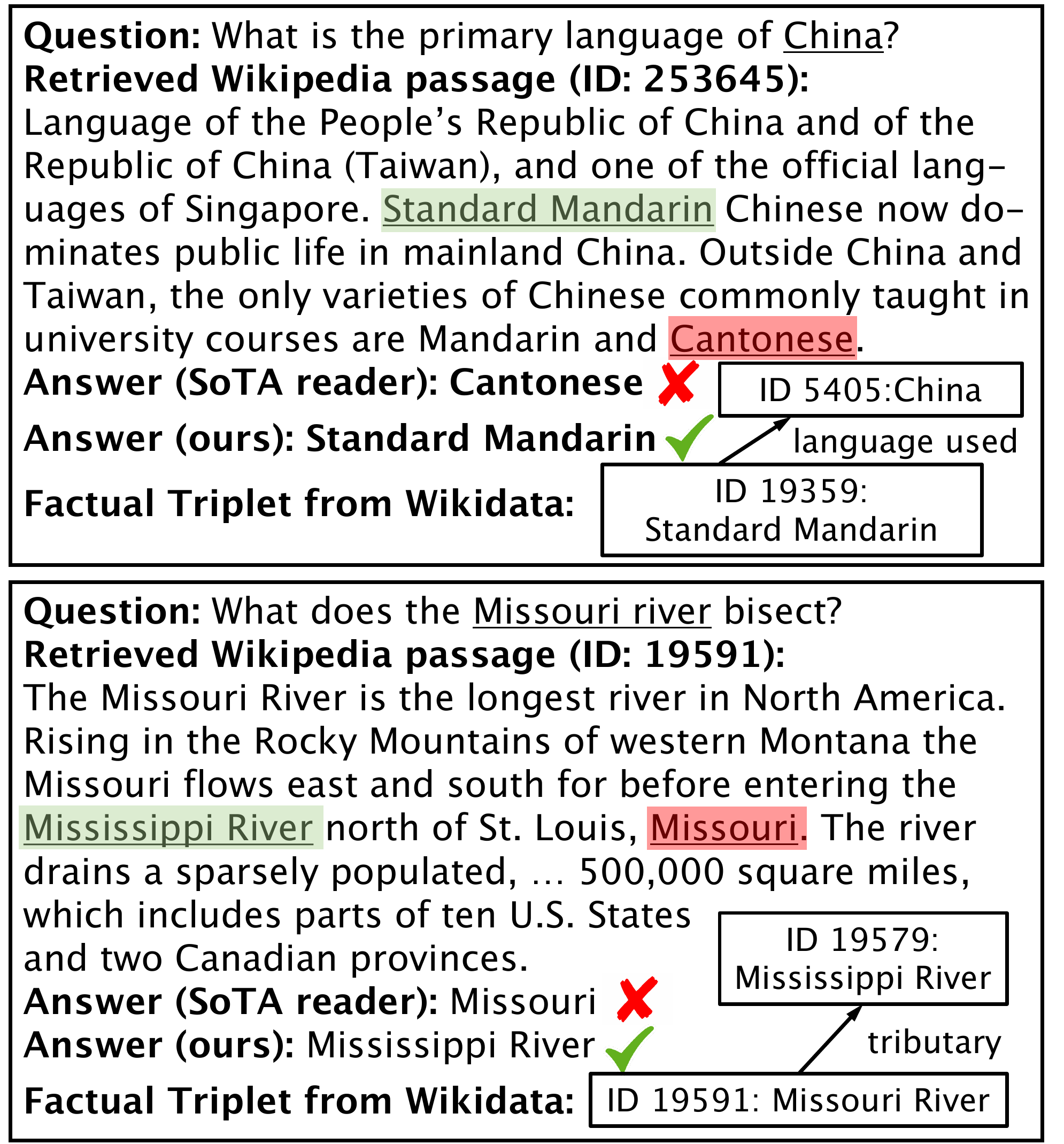}
    \vspace{-0.25in}
    \caption{The answers produced by the SoTA reader FiD contradict the facts in the knowledge graph.}
    \label{fig:triplet_eg}
\end{figure}

\blfootnote{* Equal contribution.}
Open-domain question answering (QA) tasks aim to answer questions in natural language based on large-scale unstructured passages such as Wikipedia~\cite{chen2020open,zhu2021retrieving}. 
A common thread of modern open-domain QA models employs a \textit{retriever-reader} pipeline, in which a retriever aims to retrieve a handful of relevant passages w.r.t. a given question, and a reader aims to infer a final answer from the received passages~\cite{guu2020realm,karpukhin2020dense,lewis2020retrieval,izacard2021leveraging}.
Although these methods have achieved remarkable advances on various open-domain QA benchmarks, the state-of-the-art readers, such as FiD~\cite{izacard2021leveraging}, still often produce answers that contradict the facts.
As shown in Figure \ref{fig:triplet_eg}, the FiD reader fails to produce correct answers due to inaccurate understanding of the factual evidence.
Therefore, instead of improving the retrievers to saturate the readers with higher answer coverage in the retrieved passages 
% that may already exist
~\cite{yu2021kg,oguz2020unik,yu2022generate}, in this work, we aim at improving the readers by leveraging structured factual triples from the knowledge graph (KG).

A knowledge graph, such as Wikidata~\cite{vrandevcic2014wikidata}, contains rich relational information between entities, many of which can be further mapped to corresponding mentions in questions and retrieved passages. To verify the possible improvements brought by the KG, we conduct a simple analysis to examine the percentage of related fact triples present on the KG in the data,
% questions whose answers can be enhanced by KG factual triplets, 
i.e., entities in questions are neighbors of answer entities in retrieved passages through any relation. We also wonder how many of the above examples are correctly answered by state-of-the-art readers.
Table \ref{tab:1hop-intro} shows that a great portion of examples (e.g., 58.1\% in WebQ) can be matched to related fact triplets on the KG. However, without using the KG, FiD frequently produces incorrect answers to questions on these subsets, leaving us significant room for improvement. 
Therefore, a framework that leverages not only textual information in retrieved passages but also fact triplets from the KG is urgently desired to improve reader performance.

\begin{table}[t]
\begin{center}
\scalebox{0.95}{
\begin{tabular}{l|c|c}
\toprule
% \multirow{2}{*}{Dataset} & \# fact- & Percentage & Error  \\
Dataset & Fact-related examples & Error rate  \\
\midrule
NQ & \ \ \ 736 \ (20.4\%) & 31.9\% \\
TriviaQA & 3,738 \ (33.0\%) & 17.4\% \\
WebQ & 1,181 \ (58.1\%) & 42.5\% \\
\bottomrule
\end{tabular}}
\vspace{-0.05in}
\caption{The error rate of state-of-the-art reader (i.e., FiD base) on the subset of data examples in the test set that have related fact triplets on the knowledge graph.}
\label{tab:1hop-intro}
\end{center}
\vspace{-0.15in}
\end{table}

In this paper, we propose a novel knowledge \underline{\textbf{Gra}}ph enhanced \underline{\textbf{p}}assag\underline{\textbf{e}} reader, namely \method, to improve the reader performance for open-domain QA.
Considering the enormous size of KGs and complex interweaving between entities (e.g., over 5 million entities and over 30 neighbors per entity on Wikidata), direct reasoning on the entire graph is intractable. Thus, we first construct a localized bipartite graph for each pair of question and passage, where nodes represent entities contained within them, and edges represent relationships between entities.
Then, node representations are initialized with the hidden states of the corresponding entities, extracted from the intermediate layer of the reader model.
Next, a graph neural network learns node representations with relational knowledge, and passes them back into the hidden states of the reader model.
Through this carefully curated design, \method takes into account both aspects of knowledge as a holistic framework.

To the best of our knowledge, we are the first work to leverage knowledge graphs to enhance the passage reader for open-domain QA.
Our experiments demonstrate that, given the same retriever and the same set of retrieved passages, \method can achieve superior performance on three open-domain QA benchmarks (i.e., NQ, TriviaQA, and WebQ) with up to 2.2 improvement on the exact match score over the state-of-the-art readers.
In particular, our proposed \method nearly doubles the improvement gain on the subset that can be enhanced by fact triplets on the KG.

\section{Related Work}

% \vspace{0.05in}
\paragraph{Text-based open-domain QA}
Mainstream open-domain QA models employ a \textit{retriever-reader} architecture, and recent follow-up work has mainly focused on improving the retriever or the reader~\cite{chen2020open,zhu2021retrieving}.
For the retriever, most of them split text paragraphs on Wikipedia pages into over 20 million disjoint chunks of 100 words, each of which is called a passage. Traditional methods such as TF-IDF and BM25 explore sparse retrieval strategies by matching the overlapping contents between questions and passages \cite{chen2017reading,yang2019end}. DPR \cite{karpukhin2020dense} revolutionized the field by utilizing dense contextualized vectors for passage indexing. Furthermore, other research improved the performance by better training strategies \cite{qu2021rocketqa}, passage re-ranking \cite{mao2021reader} or directly generating passages \cite{yu2022generate}. Whereas for the reader, extractive readers aimed to locate a span of words in the retrieved passages as answer~\cite{karpukhin2020dense,iyer2021reconsider,guu2020realm}. On the other hand, FiD and RAG, current state-of-the-art readers, leveraged encoder-decoder models such as T5 to generate answers \cite{lewis2020retrieval,izacard2021leveraging}. 
Nevertheless, these readers only used text corpus, failing to capture the complex relationships between entities, and hence resulting in produced answers contradicting the facts.

\paragraph{KG-enhanced methods for open-domain QA}
Recent work has explored incorporating knowledge graphs (KGs) into the \textit{retriever-reader} pipeline for open-domain QA~\cite{min2019knowledge,zhou2020knowledge,oguz2020unik,yu2021kg,hu2022empowering,yu2022survey}. For example, Unik-QA converted structured KG triples and merged unstructured text together into a unified index, so the retrieved evidence has more knowledge covered.  
% retrieved KG triplets associated with the question and converts them to text corpus. 
Graph-Retriever~\cite{min2019knowledge} and GNN-encoder~\cite{liu2022gnn} explored passage-level KG relations for better passage retrieval. KAQA \cite{zhou2020knowledge} improved passage retrieval by re-ranking according to KG relations between candidate passages. KG-FiD \cite{yu2021kg} utilized KG relations to re-rank retrieved passages by a KG fine-grained filter. However, all of these retriever-enhanced methods focused on improving the quality of retrieved passages before passing them to the reader model. So, they still suffered from factual errors. Instead, our \method is the first work to leverage knowledge graphs to enhance the reader, which is orthogonal to these existing KG-enhanced frameworks and our experiments demonstrate that with the same retriever and the same set of retrieved passages, \method can outperform the state-of-the-art reader FiD by a large margin.

\section{Proposed Method: \method}
\label{sec:method}

\begin{figure*}
    \centering
    {\includegraphics[width=1\textwidth]{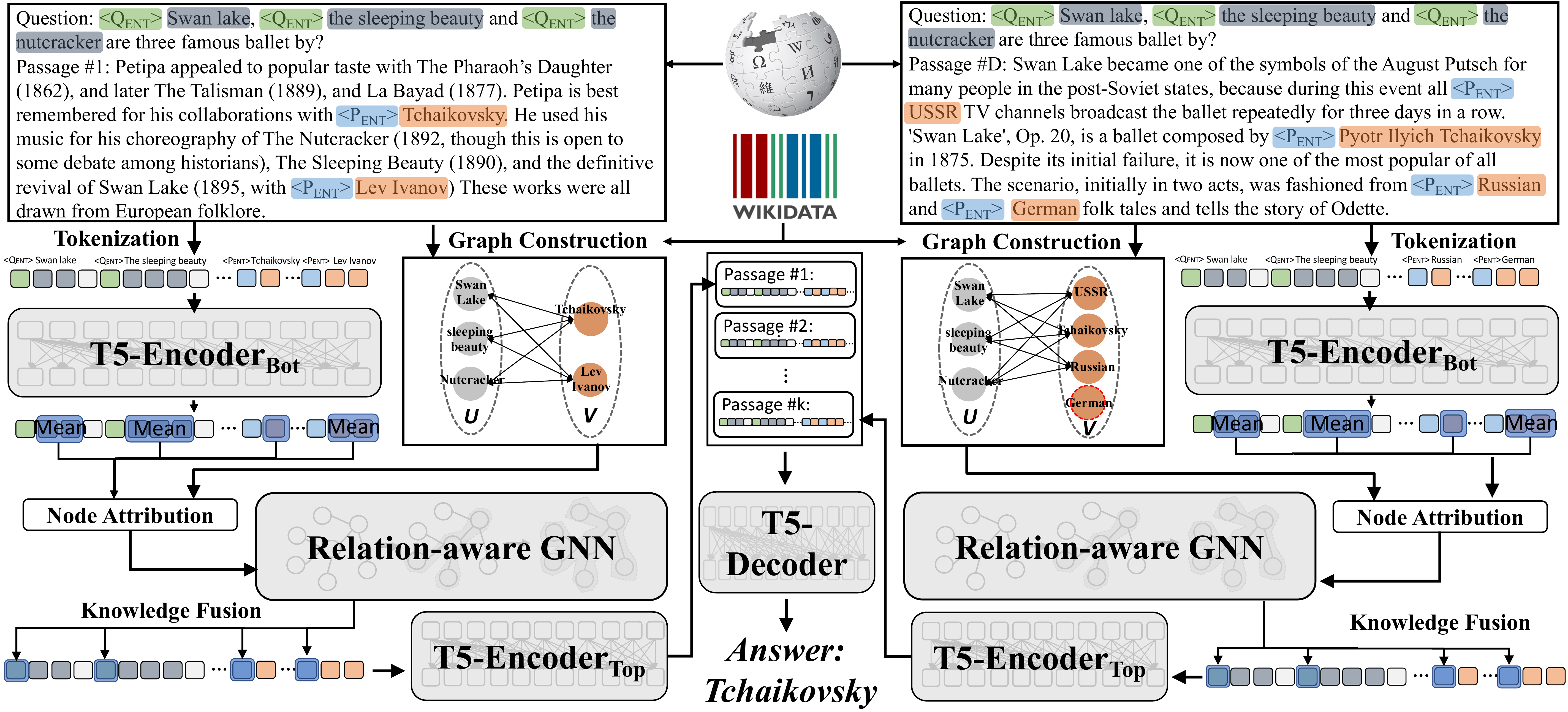}}
    \caption{Two documents are independently encoded by our \method with their corresponding localized bipartite graphs, leveraging both textual and structured information. The relation-aware GNN learns the structured knowledge from the localized bipartite graphs, attributed with entity representations extracted from the T5-Encoder\textsubscript{Bot}. The node representations are then fused into the T5-Encoder\textsubscript{Top}, which provides the hidden representations of the document. Finally, the T5-Decoder takes hidden states from all documents and generates the answer. }
    \label{fig:system}
    % \vspace{-0.1in}
\end{figure*}

In this section, we elaborate on the details of the proposed \method. Figure \ref{fig:system} shows its overall architecture. \method adopts a retriever-reader pipeline. Specifically, given a question, it first utilizes DPR to retrieve top-$k$ relevant passages from Wikipedia ($\S$\ref{sec:retrieval}). Then, to peruse the retrieved passages, it constructs a localized bipartite graph for each pair of question and passage ($\S$\ref{sec:graph}). The constructed graphs possess tractable yet rich knowledge about the facts among connected entities. Finally, with the curated graphs, structured facts are learned through a relation-aware graph neural network (GNN) and fused into token-level representations of entities in the passages ($\S$\ref{sec:gnn}). 

\subsection{Passage Retrieval}
\label{sec:retrieval}

Given a collection of $K$ passages, the goal of the retriever is to map all the passages in a low-dimensional vector, such that it can efficiently retrieve the top-$k$ passages relevant to the input question. Note that $K$ can be very large (e.g., over 20 million in our experiments) and $k$ is usually small (e.g., 100 in our experiments).

\begin{figure}[t]
\includegraphics[width=0.492\textwidth]{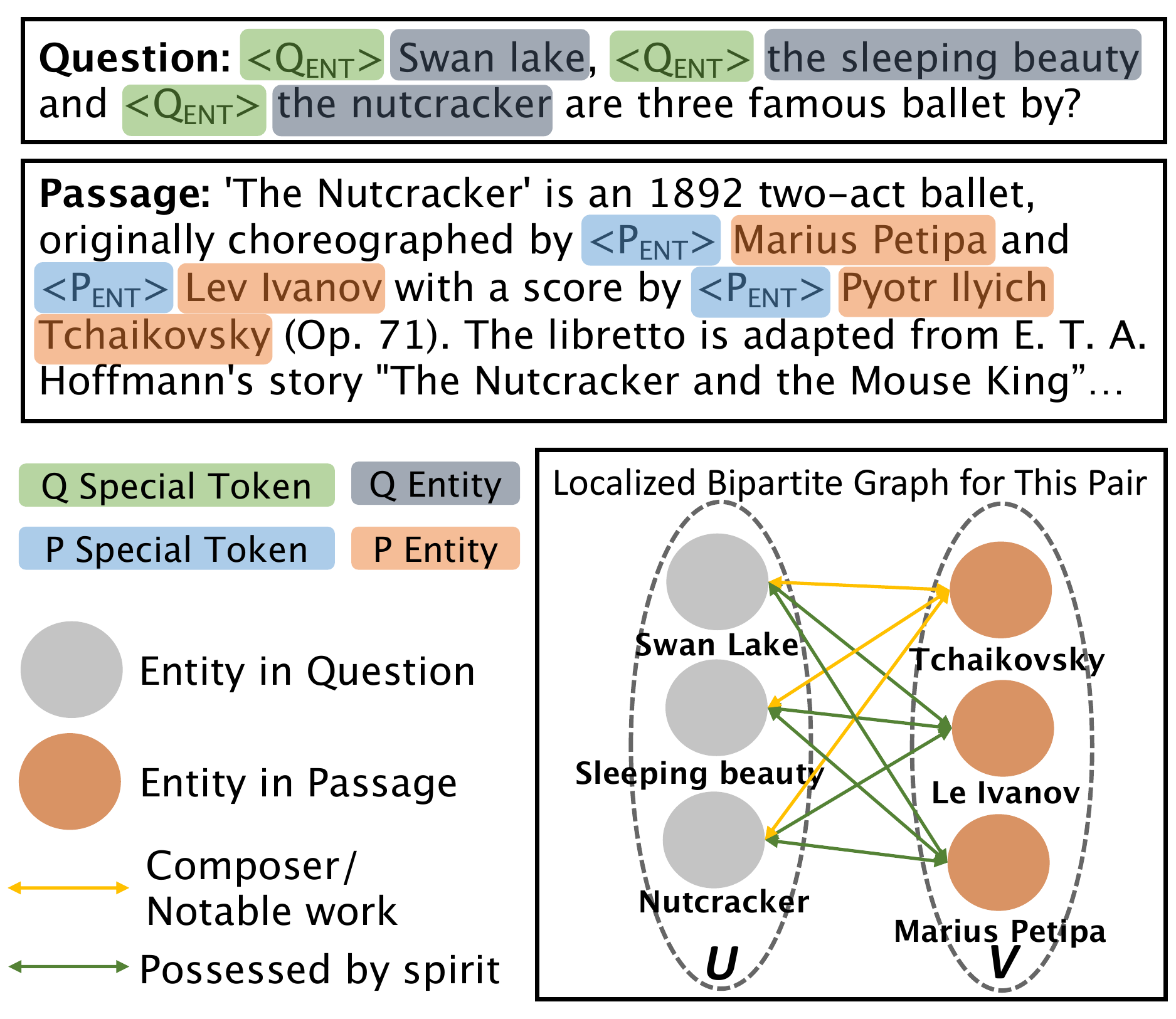}
\vspace{-0.1in}
    \caption{
    Given a pair of question and passage, the proposed \method constructs a localized bipartite graph.}
    \label{fig:qp_example}
    \vspace{-0.1in}
\end{figure}

Following DPR~\cite{karpukhin2020dense}, we employ two independent BERT~\cite{devlin2019bert} models to encode the question and the passage separately, and estimate their relevance by computing a single similarity score between their $\text{[CLS]}$ token representations. 
% DPR utilizes a bi-encoder architecture, which retrieves passages based on similarities w.r.t the question, defined as the inner product between question and passage representations created by their corresponding encoders \cite{karpukhin2020dense}. 
Specifically, given a question $q$ and a passage $p_i \in \{ p_1,p_2, ... , p_{K} \}$, we encode $q$ by a question encoder $E_Q(\cdot): q \rightarrow \mathbb{R}^d$ and encodes $p_i$ by a passage encoder $E_P(\cdot): p \rightarrow \mathbb{R}^d$, where $d$ is the hidden dimension of the used BERT. The ranking score $r_q^i$ of $p_i$ w.r.t $q$ is calculated as:
\begin{equation}
    r_q^i = E_Q(q)^\intercal \cdot E_P(p_i).
\end{equation}
We select $k$ passages whose ranking scores $r_q$ are top-$k$ highest among all $K$ passages. Before passing the retrieved passages into the reader model, we process each question and passage by inserting special tokens before each entity. For entities in each passage, we use the special token \texttt{<P\textsubscript{ENT}>}; for those in the question, we use another special token \texttt{<Q\textsubscript{ENT}>}, as shown in Figure \ref{fig:qp_example}. The special tokens play an important role in our proposed reader model, which is illustrated in more detail in $\S$\ref{sec:gnn}.

\subsection{KG-enhanced Passage Reader}

\subsubsection{Graph Construction}
\label{sec:graph}
Given the retrieved and processed passages, our proposed \method utilizes the factual triplets from KGs to construct localized bipartite graphs for each question-passage pair.
A KG is defined as a set of triplets $KG=\{(e_h, r, e_t)\}$, where $e_h$, $e_t$, and $r$ refer to a head entity, a tail entity, and a corresponding relation between them, respectively. 
Knowledge graphs represent facts in the simple format of triplets, which can easily be leveraged to enrich our knowledge. Taking the question-passage pair in Figure \ref{fig:qp_example} as an example, without any prior knowledge about the authorship of the ballets, the selection of answers between ``Marius Petipa'', ``Lev Ivanov'' and ``Tchaikovsky'' is difficult. Nonetheless, factual triplets from the KG show that these three ballets are only ``possessed by spirit'' by ``Marius Petipa'' and ``Lev Ivanov''. And their ``composer'' relations with ``Tchaikovsky'' make the answer obvious. 
By fusing such relational facts from KG triplets, the reader can better comprehend the concrete facts between involved entities and hence improve the performance for open-domain QA.

One naive solution could be fetching a sub-graph from the KG where all entities involved in the questions and the passages are included. While such design preserves all potentially relevant information, it suffers from dimensionality and noise issues. 
Therefore, we proposed to construct a localized bipartite graph for each question-passage pair, where only relational facts on relevant entities are kept. 
That is, in order to prune noisy peripheral relations, only the factual relations between question entities and passage entities are included in the localized bipartite graph.
% As an alternative, for each pair of a question and a passage, we propose a localized bipartite graph, where only useful entities are kept. 
Let a bipartite graph be denoted as $G=(\mathcal{U},\mathcal{V},\mathcal{E})$, where $\mathcal{U}$ and $\mathcal{V}$ are two disjoint sets of nodes, and $\mathcal{E}$ is the edge set containing edges that connect nodes from $\mathcal{U}$ to $\mathcal{V}$, or vice versa. Specifically, in \method, $\mathcal{U}$ and $\mathcal{V}$ are defined as the entity nodes in the question and the retrieved passage, respectively. There exists an bi-directional edge $(e_h, e_t)$ between $e_h \in \mathcal{U}$ and $e_t \in \mathcal{V}$ if and only if $\{(e_h, r_{h,t}, e_t) : r_{h,t} \in \mathcal{R}, (e_h, r_{h,t}, e_t) \text{ or } (e_t, r_{t,h}, e_h)\in KG\} \neq \varnothing$, where $\mathcal{R}$ denotes the set of all relation types on KG. 
Isolated nodes without any neighbors are removed from the graph. An example graph is shown in Figure \ref{fig:qp_example}, and Table \ref{tab:graph_stat} (in the appendix) shows the statistics of the constructed graphs. 

\subsubsection{Factual Relation Fusion}
\label{sec:gnn}
In this section, we illustrate how the proposed \method fuses structured knowledge from our constructed localized bipartite graphs into the reader. 

\method uses FiD~\cite{izacard2021leveraging} as the backbone architecture, which utilizes a T5~\cite{raffel2019exploring} for encoding and decoding. To answer a question $q$, the input consists of $k$ retrieved documents \{doc\textsubscript{1}, doc\textsubscript{2}, $\cdots$, doc\textsubscript{$k$}\}, where doc\textsubscript{i} denotes to the concatenation of the token sequence of $q$ and the token sequence of $i$-th retrieved passage $p_i$. Specifically, doc\textsubscript{i} = \{$q^1$, $q^2$, $\cdots$, $q^t$, $p_i^1$, $p_i^2$, $\cdots$, $p_i^o $\}, where $t$ and $o$ are the length of the question and the passage sequence, respectively\footnote{For the simplicity of the notation, we assume all questions and passages have the equal length $t$ and $o$, respectively.}. 
Given doc\textsubscript{i}, $G_i$ denotes the localized bipartite graph constructed from it. And $I_s(\text{doc\textsubscript{i}})$, $I_e(\text{doc\textsubscript{i}})$, and $I_t(\text{doc\textsubscript{i}})$ denote the indices of the start, end, and special tokens of all entities in doc\textsubscript{i}, respectively. 

To fuse the relational knowledge from our constructed graphs, we split the encoder $\mathrm{Enc}(\cdot):\text{doc} \rightarrow \mathbb{R}^{(t+o) \times d}$ of the reader (i.e., the encoder of T5) into two partitions $\mathrm{Enc}_{\text{top}}(\cdot)$ and $\mathrm{Enc}_{\text{bot}}(\cdot)$.
The bottom part $\mathrm{Enc}_{\text{bot}}(\cdot)$ contains the first $L$ layers of $\mathrm{Enc}(\cdot)$ and the top part $\mathrm{Enc}_{\text{bot}}(\cdot)$ contains the rest, where $L$ is a hyper-parameter.
Given doc\textsubscript{i}, $\mathrm{Enc}_{\text{bot}}(\cdot)$ delivers its encoded intermediate hidden states $\mathbf{H}_i^b \in \mathbb{R}^{(t+o) \times d}$, formulated as:
\begin{equation}
    \mathbf{H}_i^b = \mathrm{Enc}_{\text{bot}}(\text{doc\textsubscript{i}}) \label{eq:enc1}.
\end{equation}
We then extract the node attributes $\mathbf{X}_i^G \in \mathbb{R}^{|\mathcal{U}\cup\mathcal{V}| \times d}$ of its corresponding graph $G_i$ from $\mathbf{H}_i^b$ according to the span of its corresponding entity. For each entity node, its attribute vector is the average of the corresponding tokens' representations. Formally,
% \vspace{-0.08in}
\begin{equation}
\vspace{-0.05in}
    \mathbf{X}_i^G = \bigoplus_{\mathclap{\substack{\texttt{StartIdx} \in I_s(\text{doc\textsubscript{i}}) \\ \texttt{EndIdx} \in I_e(\text{doc\textsubscript{i}})}}}
                                \text{avg}\Big(\mathbf{H}_i^b[\texttt{StartIdx} \text{:} \texttt{EndIdx}]\Big)^\intercal,
     \label{eq:collect}
     \vspace{-0.05in}
\end{equation}
where $\bigoplus$ denotes the vertical concatenation. 

We use a relation-aware graph neural network (GNN) to conduct relation-aware message passing on the constructed graph $G_i$ with attributes $\mathbf{X}_i^G$, denoted as $\mathrm{GNN}(\cdot, \cdot):G \times \mathbb{R}^{|\mathcal{U}\cup\mathcal{V}| \times d} \rightarrow \mathbb{R}^{|\mathcal{U}\cup\mathcal{V}| \times d}$, and the learning process is formulated as:
\begin{equation}
\vspace{-0.05in}
    \mathbf{H}_i^G = \text{GNN}\big(G_i, \mathbf{X}_i^G\big) \label{eq:agg},
    \vspace{-0.05in}
\end{equation}
where the learned node representations $\mathbf{H}_i^G$ contain relational knowledge extracted from the KG as well as contextualized knowledge from the encoder. For the coherence of reading, the details of $\mathrm{GNN}(\cdot, \cdot)$ are described later in this subsection.

With the learned entity node representations $\mathbf{H}_i^G$ containing knowledge from the fact relations, we leverage the special tokens to fuse them back into the reader. Specifically, we have $\mathbf{H}_{i}^{u} = \mathbf{H}_{i}^{b}$ then
\begin{equation}
    \mathbf{H}_{i}^{u}[I_t(\text{doc\textsubscript{i}})] = \mathbf{H}_{i}^{b}[I_t(\text{doc\textsubscript{i}})] + \mathbf{H}_i^G, \label{eq:att}
\end{equation}
where $[\cdot]$ is the indexing operation. The updated contextualized representations $\mathbf{H}_{i}^{u}$ are then used as the input of the top part of the encoder to enable further information exchanges among regular tokens and the updated special tokens:
\begin{equation}
    \mathbf{H}_i = \mathrm{Enc}_{\text{top}}(\mathbf{H}_{i}^{u}). \label{eq:enc2}
\end{equation}
Given the question $q$, \method forwards all $k$ retrieved documents through the above-described encoding process, and acquires the hidden states of all documents 
$\{\mathbf{H}_i\}_{i=1}^{k}$.
% $\{ \mathbf{H}_i : \text{ i} \in \{1 ,2,...,k\} \}$.
These hidden states are then concatenated and sent to the decoder $\mathrm{Dec}(\cdot)$ for answer generation. Formally,
% \vspace{-0.1in}
\begin{equation}
\vspace{-0.05in}
    \text{answer} = \mathrm{Dec}\big(\bigoplus_{i=0}^{k} \mathbf{H}_i\big). \label{eq:dec}
    \vspace{-0.05in}
\end{equation}
To sum up, the workflow of our proposed \method can be concluded as the following four steps: (i) get the initial contextualized representations via $\mathrm{Enc}_{\text{bot}}$ (Equation (\ref{eq:enc1})) and the node attributes (Equation (\ref{eq:collect})), (ii) fuse fact relation  by a relation-aware GNN (Equations (\ref{eq:agg}) and (\ref{eq:att})), (iii) exchange additional information via $\mathrm{Enc}_{\text{top}}$ (Equation (\ref{eq:enc2})), and (iv) generate the answer by the decoder (Equation (\ref{eq:dec})).

\paragraph{Relation-aware GNN}
Here we elaborate the details of the aforementioned $\text{GNN}(\cdot,\cdot)$. Typically, each GNN layer~\cite{hamilton2017inductive,kipf2016semi} can be formulated as 
\begin{equation}
\scale[0.92]{
    \mathbf{h}_v^{(n)} = \text{AGG}^{(n)}\Big(\{ \text{TRANS}^{(n)} \big( \mathbf{h}_u^{(n-1)}\big)  : u \in \mathcal{N}(v) \}\Big),}
\end{equation}
where $\mathcal{N}(v)$ is the set of neighbors for node $v$ including itself, $n$ is the index of the current layer, and $\mathbf{h}_v^{n}$ denotes the representation of node $v$ at the $n$-th layer. 
The transform function $\text{TRANS}(\cdot)$ projects node representations from the previous layer to a new vector space for message passing. The aggregation function $\text{AGG}(\cdot)$ takes a set of node representations and aggregates them as a vector in a unified view \cite{kipf2016semi,velivckovic2018graph,zhang2019heterogeneous,fan2022heterogeneous,ju2022adaptive}. 
Our proposed \method uses a multi-layer perceptron as $\text{TRANS}(\cdot)$ in each layer. That is,
\begin{equation}
\vspace{-0.05in}
    \mathbf{A}^{(n)} = \sigma\Big(\mathbf{H}^{(n-1)} \cdot \mathbf{W}_t^{(n)} + \mathbf{b}_t^{(n)}\Big) 
    \vspace{-0.05in}
\end{equation}
where $\mathbf{A}^{(n)}$ is the intermediate embedding to be used by $\text{AGG}(\cdot)$, $\mathbf{H}^{(0)} = \mathbf{X}_i^{G}$ as aforementioned in Equation~(\ref{eq:collect}), $\mathbf{W}_t^{(n)} \in \mathbb{R}^{d \times d}$ and $\mathbf{b}_t^{(n)} \in \mathbb{R}^d$ denote the learnable parameters, and $\sigma(\cdot)$ refers to the non-linear activation function.

For the aggregation function $\text{AGG}(\cdot)$, we explore a relation-aware attention mechanism. Different from GAT~\cite{velivckovic2018graph} that considers only node representations for the edge attention weight, \method also incorporates the relation representations between nodes. At layer $n$, for each node $v$, its representation $\mathbf{h}_v^{(n)}$ is calculated by
\begin{align}
\vspace{-0.1in}
\begin{split}
    & e_{v,u}^{(n)} = \Big(\mathbf{a}_v^{(n)} \scaleobj{0.85}{\bigoplus} \text{avg}\big(\mathrm{Enc}(r_{v,u})\big) \scaleobj{0.85}{\bigoplus} \mathbf{a}_u^{(n)}\Big) \cdot \mathbf{W}_e^{(n)}, \\
    & \alpha_{v,u}  = \frac{e_{v,u}^{(n)}}{\sum_{m \in \mathcal{N}(v)} e_{v,m}^{(n)}}, \; \mathbf{h}_v^{(n+1)} = \sum_{\mathclap{u \in \mathcal{N}(v)}} \alpha_{v,u} \cdot \mathbf{a}_v^{(n)}, \label{eq:gnn}
\end{split}
\vspace{-0.1in}
\end{align}
where $\mathbf{a}_v^{(n)}$ is the node $v$'s representation in $\mathbf{A}^{(n)}$, $\mathbf{W}_e^{(n)} \in \mathbb{R}^{3d \times 1}$ calculates the importance score of node $u$ to node $v$, considering contextualized representations of the connected two nodes and language model's understanding of their relationship (i.e., avg($\mathrm{Enc}(r_{u,v})$)\footnote{In our implementation, we calculate and buffer $\{\text{avg}(\mathrm{Enc}(r)): r \in R\}$ before every batched forward and simply query the representation of any relation as needed.}. We further extend this schema to the multi-head attention pipeline by having multiple operations as described in Equation (\ref{eq:gnn}) running in parallel. That is,
\begin{equation}
\vspace{-0.05in}
    \mathbf{H}^{(n)} = \sum\nolimits_{m=0}^{M} \mathbf{H}^{(n, m)},
    \vspace{-0.05in}
\end{equation}
where $M$ denotes the number of heads, and $\mathbf{H}^{(n,m)}$ refers to the learned representations of the $m$-th head at $n$-th layer. Finally, the node representations $\mathbf{H}_i^{(N)}$ are used as the output of $\text{GNN}(\cdot, \cdot)$, where $N$ is the number of layers in the GNN. 

In summary, our relation-aware GNN combines the current reader's understanding of the factual relationships among nodes (i.e., avg($\mathrm{Enc}(r)$) with the intermediate hidden states $\mathbf{X}_i^G$ from $\mathrm{Enc}_{\text{bot}}(\cdot)$. 
% In this process, higher attention weights are assigned to edges that promote factual relationships about answers, which further encourages message passing between reasonable answer candidates and question entities. 
Enriched by structured fact relations, entity node representations are then fused back into the reader's encoder so that our \method can comprehend facts between entities during the encoding process. 

\section{Experiments}

In this section, we conduct comprehensive experiments on three community-acknowledged public open-domain QA benchmarks: Natural Questions (NQ) based on Google search queries, TriviaQA based on questions from trivia and quiz-league websites, and Web Questions (WebQ) based on questions from Google Suggest API \cite{kwiatkowski2019natural,joshi2017triviaqa,berant2013semantic}. We explore the same train / dev / test splits and preprocessing techniques as used by \cite{izacard2021leveraging,karpukhin2020dense}. 

\subsection{Experimental Setup}

\paragraph{Retrieval Corpus} 
We followed the same process as used in \citep{karpukhin2020dense,lewis2020retrieval} for preprocessing Wikipedia pages.
We split each Wikipedia page into disjoint 100-word passages, resulting in 21 million passages in total. As for the knowledge graph used to construct our localized bipartite graphs, we used English Wikidata~\cite{vrandevcic2014wikidata}. The total number of aligned entities, relations, and triplets on Wikidata is 2.7M, 630, and 14M respectively\footnote{The Wikipedia and Wikidata were all collected in December of 2019. We only used the most visited top 1M entities.}. 
We used ELQ \cite{li2020efficient} to identify mentions in the question and retrieved passages, and link them to corresponding entities on Wikidata. 
\begin{table*}
\centering
\setlength{\tabcolsep}{4mm}{
\begin{tabular}{c|lcccc}
\toprule
Group & Model & \#params & NQ & TriviaQA & WebQ \\  
\cmidrule(r){1-6}
\multirow{2}{*}{(i)}& T5-11B & 11B & 32.6 & 42.3 & 37.2 \\
& GPT-3 (64-shot) & 175B & 29.9 & - & 41.5 \\
\cmidrule(r){1-6}
\multirow{3}{*}{(ii)}& DPR  
& 110M & 41.5 & 56.8 & 41.1$^{\star}$ \\
& RIDER 
& 626M & 48.3 & - & -\\
& RECONSIDER  
& 670M & 45.5 & 61.7 & -\\ \cmidrule(r){1-6}
\multirow{3}{*}{(iii)}& Graph-Retriever%
& 110M & 34.7 & 55.8 & 36.4\\ 
& Path-Retriever%
& 445M & 31.7 & - & -\\
& KAQA%
& 110M  & - & 64.1 & -\\ \cmidrule(r){1-6}
\multirow{5}{*}{(iv)}& REALM 
& 330M &  40.4 & - & 40.7 \\
& RAG 
& 626M &  44.5 & 56.1 & 45.2$^{\star}$ \\
& Joint Top-K 
& 990M & 48.1 & 59.6 & -\\
\cmidrule(r){2-6}
& FiD (base)%
& 440M & 48.2 & 65.0 & 46.5     \\ 
& FiD (large)%
& 990M & 51.4 & 67.6 & 50.5 \\ \cmidrule(r){1-6}
\multirow{2}{*}{(v)}& \method (base)
& 454M & 48.7 (0.5$\uparrow$)  & 66.2 (1.2$\uparrow$)& 48.1 (1.6$\uparrow$) \\
& \method (large) & 1.01B & \textbf{53.5} (2.1$\uparrow$) & \textbf{69.8} (2.2$\uparrow$) & \textbf{51.7} (1.2$\uparrow$)\\ 
\bottomrule
\end{tabular}}
\vspace{-0.1in}
\caption{Exact match scores over the test sets of Natural Questions, TriviaQA and Web Questions. We put the training details (such as learning rate, batch size, dev performance, etc) corresponding to the performance of our \method in Table \ref{tab:computing}. Numbers in parenthesis are improvements of \method over the corresponding best-performing baseline. Note that $\star$ means model is warmed with external training data from Natural Questions.}
\vspace{-0.1in}
\label{tab:main}
\end{table*}
% \vspace{-0.05in}
\paragraph{Implementation Details} 
In \method, involved hyper-parameters are the number of retrieved passages $k$, the number of GNN layers $N$, the number of GNN head $M$, and the encoder layer index $L$, where $\mathrm{Enc}_{\text{top}}$ and $\mathrm{Enc}_{\text{bot}}$ are partitioned). We explore $k=100$, $N=2$, $L=3$ and $M=8$ as the default setup. The hidden dimension of the GNN in \method are set to the dimension of its language model (i.e., $d=768$ for the base configuration and $d=1024$ for the large configuration). Since $N=2$ and $M=8$ are the standard values for most GNNs with the attention mechanism, able to capture 2-hop neighbor information while being stable \cite{velivckovic2018graph}, we simply follow the same principle. 

For the encoder layer index $L$, we search the optimal value in the range of $\{3, 4, 6, 8, 9\}$. According to our experiment, we observe that different selections of $L$ don't have much impact on the performance, indicating that the infusion of structured knowledge doesn't correlate with the contextualization of the entity embedding. However, we do observe a faster convergence rate for lower $L$ values. So for faster convergence, we set $L$ to 3. Other hyper-parameter selections related to training with best performance across all datasets are shown in Table \ref{tab:computing} in $\S$\ref{sec:package}. The software and hardware information can be found in $\S$\ref{sec:package} in the appendix. 

\paragraph{Evaluation Metrics} We use the standard evaluation metric for open-domain QA: exact match score (EM) \cite{rajpurkar2016squad,zhu2021retrieving}. An answer is considered correct if and only if its normalized form\footnote{We use the same normalization procedure as introduced in \cite{karpukhin2020dense}.} has a match in the acceptable answer list. For all experiments, we conduct 3 runs with different random seeds and report the average. 
% \vspace{-0.05in}
\subsection{Baseline Models}
We compare \method with four groups of baselines: (i) The first group includes closed-book models, where no Wikipedia document is provided during training and inference: T5-11B \cite{raffel2019exploring} and GPT-3 \cite{brown2020language}. (ii) The second contains extractive models, which utilize passages extracted by enhanced retrievers and find the span of the answer: DPR \cite{karpukhin2020dense}, RIDER \cite{mao2021reader}, RECONSIDER \cite{iyer2021reconsider}. (iii) The third includes approaches that utilize KG for retrieving: Graph-Retriever \cite{min2019knowledge}, Path-Retriever \cite{asai2019learning}, and KAQA \cite{zhou2020knowledge}. (iv) The last group contains advanced generative readers: RAG \cite{lewis2020retrieval}, REALM \cite{guu2020realm}, Top-K \cite{sachan2021end} and FiD~\cite{izacard2021leveraging}. For FiD, we compare both its base and large versions, and for other baselines we compare their best-performing versions.

We note that our method is the first work using the knowledge graph to improve the reader performance for open-domain QA. Hence, this is orthogonal to existing works using the knowledge graph to improve passage retrieval~\cite{liu2022gnn} or re-ranking~\cite{yu2021kg}. Our experiments show that with the same retriever and the same set of retrieved passages, \method can outperform the state-of-the-art reader FiD by a large margin.

\subsection{Experimental Results}

\subsubsection{Comparison with Baselines}
Table \ref{tab:main} shows the model performance of 13 baselines as well as our \method. We can observe that our proposed \method can significantly outperform the best performing baselines across all datasets over both base and large configurations. Specifically, \method improves FiD by 0.5, 1.2, and 1.6 EM score on the base model, and 2.1, 2.2, and 1.2 EM scores on the large model on NQ, TriviaQA and WebQ, respectively. Albeit being competitive on all datasets, \method brings more improvements on TriviaQA and WebQ than NQ. We believe the reason is that on NQ, the percentage of questions favorable from factual KG relations among all questions, as shown in Table \ref{tab:1hop-intro} and Table \ref{tab:solvable}, is relatively lower, compared to TriviaQA and WebQ.
On the large configuration, even though some questions are not directly favored by structured facts, the additional information re-routing from \method still benefits with additional learning capability, which outperfoms FiD large for 2.1 EM score. We also note that the performance of RAG on WebQ is better than FiD base and close to \method base, which is caused by a tremendous amount of additional training on other open-domain QA datasets.

\subsubsection{Ablation Study}
We design two variants for our \method. The first is \method without considering relations between entity nodes, i.e., $ \text{avg}(\mathrm{Enc}(r_{u,v}))$ in Equation (\ref{eq:gnn}) is deleted. The goal is to validate the improvements brought by relational knowledge, which is denoted as w/o Rel in Table ~\ref{tab:ablation}. The second is \method without considering the relations as well as neighbor differences. The goal is to validate if \method can differentiate the important neighbor without the attention mechanism, which is denoted as w/o Att. As shown in Table \ref{tab:ablation}, we can observe that the performance drops when removing any of the two mechanisms, demonstrating their effectiveness and further validating the rich inductive bias brought by the factual relations between entities. Besides, we also notice that the incorporation of relation is more important than the attention mechanism.

\begin{table}[t]
\begin{center}

\setlength{\tabcolsep}{1.5mm}{
\scalebox{0.98}{\begin{tabular}{l|cccc}
\toprule
% \multirow{2}{*}{Dataset}& \multicolumn{4}{c}{EM} \\
Datasets & FiD & \method & w/o Rel & w/o Att \\ \cmidrule(r){1-5}
\multirow{2}{*}{NQ}
& 48.2 & \textbf{48.7} & 48.6 & 48.3 \\
& 51.4 & \textbf{53.5} & 53.4 & 53.1 \\ 
\cmidrule(r){1-5}
\multirow{2}{*}{TriviaQA}
& 65.0 & \textbf{66.2} & 65.7 & 65.8 \\
& 67.6 & \textbf{69.8} & 69.5 & 69.6 \\ 
\cmidrule(r){1-5}
\multirow{2}{*}{WebQ}
& 46.5 & 48.1 & \textbf{48.4} & 48.2  \\
& 50.5 & \textbf{51.7} & 51.5 & 51.0 \\ 
\bottomrule
\end{tabular}}}
\vspace{-0.05in}
\caption{Ablation study of \method without relation knowledge or attention mechanism. The first line refers to the performance on the base model; whereas the second line refers to the large model.}
\label{tab:ablation}
\end{center}
\end{table}

\begin{table}[t]
\begin{center}
\setlength{\tabcolsep}{1.5mm}{
\scalebox{0.95}{\begin{tabular}{l|cc|cc}
\toprule
\multirow{2}{*}{Dataset} & \multirow{2}{*}{Subset\%}& Model& \multicolumn{2}{c}{EM among subset} \\
% \cmidrule(lr){4-5}
 & & size & \ \ \ FiD & \method \\
\cmidrule(r){1-5}
\multirow{2}{*}{NQ}
& \multirow{2}{*}{20.3\%} & base & \ \ \ 68.1 & \textbf{70.7} (2.6$\uparrow$) \\
& & large & \ \ \ 69.3 & \textbf{71.9} (2.6$\uparrow$) \\ 
\midrule
\multirow{2}{*}{TriviaQA}
& \multirow{2}{*}{33.0\%} & base & \ \ \ 82.6 & \textbf{86.4} (3.8$\uparrow$) \\
& & large & \ \ \ 85.6 & \textbf{88.9} (3.3$\uparrow$) \\
\midrule
\multirow{2}{*}{WebQ}
& \multirow{2}{*}{58.1\%} & base & \ \ \ 57.5 & \textbf{61.2} (3.7$\uparrow$) \\
& & large & \ \ \ 62.7 & \textbf{65.1} (2.4$\uparrow$) \\
\bottomrule
\end{tabular}}}
\caption{Exact match score on the subset of questions that can be enhanced by factual triplets from the KG.}
\label{tab:solvable}
\vspace{-0.2in}
\end{center}
\end{table}

\begin{figure}[t]
\vspace{-0.15in}
\includegraphics[width=0.5\textwidth]{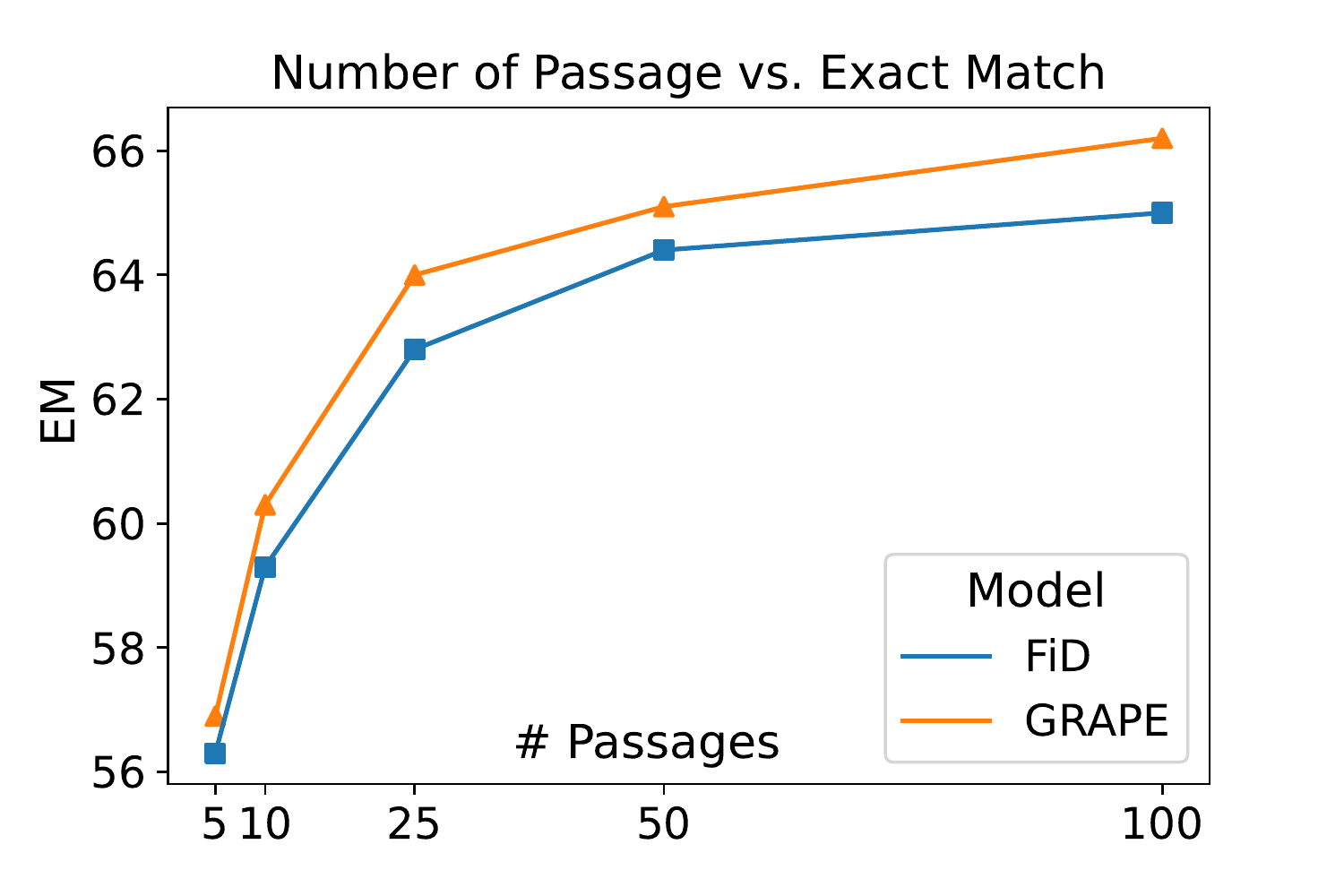}
\vspace{-0.3in}
    \caption{The performance on the test set of TriviaQA w.r.t. the number of passages.}
    \label{fig:em_passage}
    \vspace{-0.15in}
\end{figure}

\subsubsection{Improvement Analysis by KG Relation}
Since \method utilizes the factual relational knowledge from KG, to validate the legitimacy of our assumption, we analyze the performance gain on the subset of questions that can be directly solved by a factual triplet on KG (i.e., constructed graphs for these questions contain at least one edge that links the answer entity to entities in questions). From Table \ref{tab:solvable}, we observe that \method significantly improves performance on this subset that factual relations from KG naturally favors. For example, on the base model, \method improves the overall performance by 0.5, 1.2, and 1.6 EM score respectively on these three datasets, and almost doubles the performance margin (i.e., 2.6, 3.8, and 3.7) on their subsets. This phenomenon demonstrates that \method tends to utilize the inductive bias we introduce through graphs and the major performance gain can be rooted in the factual relational knowledge from KG, which further validates the legitimacy of \method.% We also include supplementary case studies in Figure \ref{fig:case} in $\S$\ref{sec:case}.

\subsubsection{Scaling with Number of Passages}

We further evaluate the performance of \method with respect to the different numbers of retrieved passages (i.e., $D = \{5, 10, 25, 50, 100\}$), as shown in Figure \ref{fig:em_passage}. We observe that given the same number of passages, \method consistently outperforms FiD, with greater performance gains given more passages. Specifically, \method performs on par with FiD with only the half amount of retrieved passages, starting from 25 retrieved passages. This phenomenon demonstrates that, when the answer is well presented in the retrieved passages, facts introduced by our curated graphs constructed from KG significantly help the reader answer questions. 
\begin{figure*}[t]
\includegraphics[width=1\textwidth]{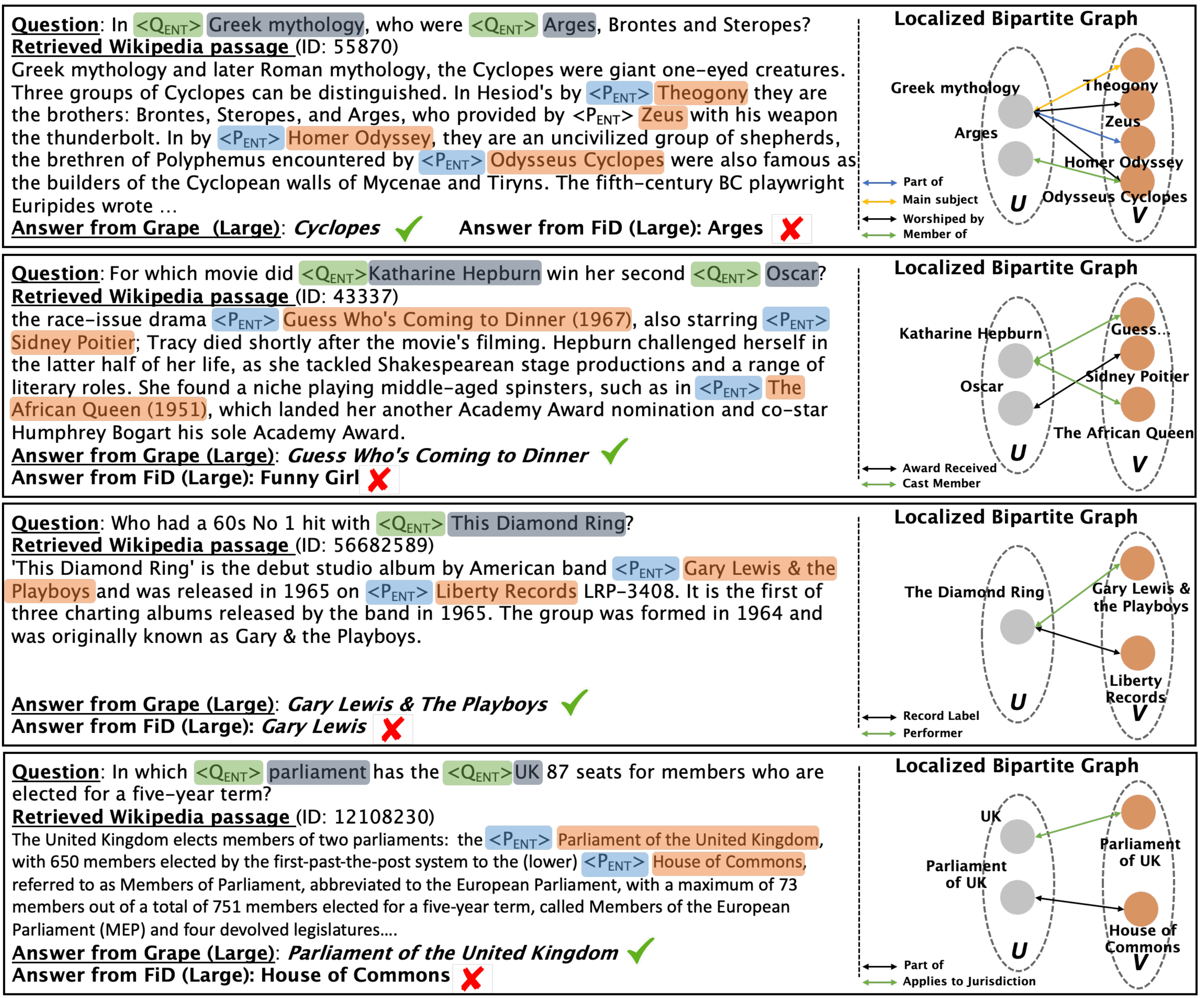}
    \vspace{-0.25in}
    \caption{Case studies on samples that are incorrectly answered by FiD but correctly answered by \method. Relations in green arrow indicate the factual relations from KG that enhance the question answering. }
    \label{fig:case}
    \vspace{-0.1in}
\end{figure*}
\subsubsection{Case Studies on KG Relations}
To further validate the improvement gain induced by \method, we analyze samples that are incorrectly answered by FiD but correctly answered by \method, and visualize the constructed graphs for these samples, as shown in Figure \ref{fig:case}. From these samples, we can observe the performance gain indeed comes from the strong enhancement brought by fact relations from their constructed graphs. For example, in the first example, with the fact relations, \method understands that ``Arges'' is a member of ``Cyclopes'', which perfectly enhances answering for the given problem. In the the third example, we can observe that FiD delivers a answer that is only partially correct. Whereas, enhanced by fact relation from KG, \method correctly answers the question, because of the triplet (``UK'', ``applies to jurisdiction'', ``Parliament of UK''). This design is tractable yet effective. Because only entities highly correlated with useful facts will be included. Specifically, passage entities unrelated to question entities are very likely to be marginal and hence removed, and only the factual triplets helpful for answering the problems are kept. Relations within passage entities are most likely peripheral and hence neglected in the bipartite graph.
\vspace{-0.05in}
\section{Conclusion}
\vspace{-0.05in}
In this work, we study the problem of open-domain QA. We discover that state-of-the-art readers fail to capture the complex relationships between entities appearing in questions and retrieved passages, resulting in produced answers that contradict the facts. To this end, we propose a novel knowledge Graph enhanced passage Reader (\method) to improve the reader performance for open-domain QA. Specifically, for each pair of question and retrieved passage, we construct an informative localized bipartite graph and explore an expressive relation-aware GNN to learn entity representations that contain contextual knowledge from passages as well as fact relations from the KG. Experiments on three open-domain QA benchmarks show that \method significantly outperforms state-of-the-art readers by up to 2.2 exact match score. In the future, we plan to enrich the structured information contained in our graphs from other external resources. 
\vspace{-0.05in}
\section{Limitations}
\vspace{-0.05in}
\method only solves errors from fact-related examples.
Besides, \method explores fact relations from Wikidata, and hence we might omit fact relations from other sources such as Freebase. 

\section*{Acknowledgement}
We appreciate Neil Shah and Yozen Liu from Snap Inc. and Zhihan Zhang from University of Notre Dame for valuable discussions and suggestions. This work is partially supported by the NSF under grants IIS-2209814, IIS-2203262, IIS-2214376, IIS-2217239, OAC-2218762, CNS-2203261, CNS-2122631, CMMI-2146076, and the NIJ 2018-75-CX-0032. Any opinions, findings, and conclusions or recommendations expressed in this material are those of the authors and do not necessarily reflect the views of any funding agencies.

\balance
\bibliography{anthology}
\bibliographystyle{acl_natbib}

\clearpage
\appendix
\section{Supplementary Appendix}
\setlength{\tabcolsep}{3.2mm}\begin{table*}[htb]
\centering
\begin{tabular}{l|ccc|ccc}
\toprule
\toprule
{\multirow{2}*{Methods}} & \multicolumn{3}{c|}{{\method (base)}} & \multicolumn{3}{c}{{\method (large)}} \\
& NQ & TriviaQA & WebQ & NQ & TriviaQA & WebQ \\  
\midrule
% Number of parameters & \multicolumn{3}{c|}{{454M}} & \multicolumn{3}{c}{{1.01B}} \\
Computing resources & \multicolumn{3}{c|}{{8x32GB Nvidia Tesla V100}} & \multicolumn{3}{c}{{8x32GB Nvidia Tesla V100}}\\
Peak memory cost & \multicolumn{3}{c|}{{Around 30.08GB (94\%)}} & \multicolumn{3}{c}{{Around 26.56GB (83\%)}} \\
Peak learning rate & 1e-4  & 6e-5  & 1e-4  & 3e-5  & 3e-5 & 6e-5 \\
learning optimizer & \multicolumn{3}{c|}{{AdamW with 2,000 warmup}} & \multicolumn{3}{c}{{AdamW with 2,000 warmup}} \\
Batch size (per device) & 3 & 3 & 3 & 1 & 1 & 1 \\
Total training steps & 50K & 30K & 20K & 50K & 30K & 20K \\
Best dev iterations & 19,500 & 16,500 & 20,000 & 45,000 & 30,000 & 27,500 \\
Best dev performance & 48.17 & 65.67 & 51.00 & 50.73 & 69.46 & 57.00 \\
\bottomrule
\bottomrule
\end{tabular}
\vspace{-0.1in}
\caption{Best training hyper-parameters for results reported in Table \ref{tab:main}. }
\label{tab:computing}
\begin{center}
\setlength{\tabcolsep}{1.5mm}{
\scalebox{1}{\begin{tabular}{lc|cccc}
\toprule
\toprule
Dataset & Split & \# Graphs per Q & \# Q Entity Node & \# P Entity Node & \# Nodes per Graph\\\cmidrule(r){1-6}
\multirow{3}{*}{NQ}
& Train & 29.5 $\pm$ 25.9 & 1.2 $\pm$ 0.4 & 2.7 $\pm$ 2.8 & 3.9 $\pm$ 3.0 \\
& Dev & 29.0 $\pm$ 25.8 & 1.2 $\pm$ 0.4 & 2.7 $\pm$2.8 & 3.8 $\pm$ 2.9 \\ 
& Test & 28.4 $\pm$ 26.0 & 1.2 $\pm$ 0.5 & 2.7 $\pm$ 2.9 & 3.9 $\pm $3.0 \\
\cmidrule(r){1-6}
\multirow{3}{*}{TriviaQA}
& Train & 34.8 $\pm$ 28.3 & 1.4 $\pm$ 0.7 & 2.7 $\pm$ 2.8 & 4.1 $\pm$ 3.1 \\
& Dev & 34.6 $\pm$ 28.7 & 1.3 $\pm$ 0.7 & 2.8 $\pm$ 2.8 & 4.2 $\pm$ 3.0 \\
& Test & 34.0 $\pm$ 28.3 & 1.4 $\pm$ 0.7 & 2.7 $\pm$ 2.8 & 4.2 $\pm$ 3.1 \\
\cmidrule(r){1-6}
\multirow{3}{*}{WebQ}
& Train & 42.2 $\pm$ 27.8  & 1.1 $\pm$0.3 & 2.8 $\pm$ 2.7 & 3.9 $\pm$ 2.8 \\
& Dev   & 43.9 $\pm$ 27.9  & 1.1 $\pm$0.4 & 2.9 $\pm$ 2.9 & 4.0 $\pm$ 3.0 \\
& Test  & 41.2 $\pm$ 26.9  & 1.1 $\pm$0.3 & 2.7 $\pm$ 2.7 & 3.8 $\pm$ 2.8 \\
\bottomrule
\bottomrule
\end{tabular}}}
\vspace{-0.1in}
\caption{Statistics of our constructed graphs for all datasets. Mean and standard deviation are calculated for each attribute. 100 passages are retrieved for each questions.}
\label{tab:graph_stat}
\end{center}
\end{table*}

% \begin{figure*}[t]
% % \vspace{-0.1in}
% \includegraphics[width=1\textwidth]{figures/case_study.pdf}
% \vspace{-0.3in}
%     \caption{Case studies on samples that are incorrectly answered by FiD but correctly answered by \method. Relations in green arrow indicate the factual relations from KG that enhance the question answering. }
%     % \vspace{-0.2in}
%     \label{fig:case}
% \end{figure*}

\subsection{Software and Hardware Environment}
\label{sec:package}
The software that \method uses includes Huggingface Transformers 4.18.0
% \footnote{\url{https://huggingface.co/docs/transformers}}
, Deep Graph Library (DGL) 0.8.2
% \footnote{\url{https://www.dgl.ai}}
and PyTorch 1.11.0
% \footnote{\url{https://pytorch.org}}
. And all our experiments are conducted on servers with 8 Tesla V100 32GB GPUs, with the RAM occupation less than 50GB. On average, updating \method's parameters costs 72 GPU hours for 10K training steps at the base configuration and 170 GPU hours at the large configuration, including intermediate validations, under the settings reported in Table \ref{tab:computing}.

\subsection{Hyper-parameter Sensitivity}
\label{sec:hyper}
In \method, hyper-parameters related to our relation-aware GNN are the number of retrieved passages $k$, the number of GNN layers $N$, the number of GNN head $M$, and encoder layer index $L$, where $\mathrm{Enc}_{\text{top}}$ and $\mathrm{Enc}_{\text{bot}}$ are partitioned). We explore $k=100$, $N=2$, $L=3$ and $M=8$ as the default setup. The hidden dimension of the GNN in \method are set to the dimension of its language model. $N=2$ and $M=8$ are the standard values for most GNNs with attention mechanism, able to capture 2-hop neighbor information while being stable \cite{velivckovic2018graph}; and we simply follow the same principle. For the encoder layer index $L$, we search the optimal value in the range of $\{3, 4, 6, 8, 9\}$. According to our experiment, we observe that different selection of $L$ doesn't have much impact on the performance, indicating that the infusion of structured knowledge doesn't rely much on the contextualization of entity embedding. However, we do observe a faster convergence rate for lower $L$ values. So for faster convergence, we set $L$ to 3.  Other hyper-parameter selections related to training with best performance across all datasets are shown in Table \ref{tab:computing}. 

\subsection{Dataset and Graph Statistics}
We explore the same training/dev/testing split as used by FiD and RAG \cite{izacard2021leveraging,lewis2020retrieval}. We further analyze the geometry of our constructed graphs for each pair of question and its corresponding retrieved passage, from the perspective of the number of available graphs, average node count and node degree, shown in Table \ref{tab:graph_stat}. We notice that the number of constructed graphs per question is proportional to the percentage of questions favorable from factual KG triplets, and further proportional to the performance gain introduced by \method. The statistics of our graphs align with the source of explored datasets: TriviaQA and WebQ are more entity-focused compared with NQ. Besides the number of graphs per question, we do not observe other difference in graphs among these three datasets. 

% \subsection{Case Studies}
% \label{sec:case}
% To further validate the improvement gain induced by \method, we analyze samples that are incorrectly answered by FiD but correctly answered by \method, and visualize the constructed graphs for these samples, as shown in Figure \ref{fig:case}. From these samples, we can observe the performance gain indeed comes from the strong enhancement brought by fact relations from their constructed graphs. For example, in the first example, with the fact relations, \method understands that ``Arges'' is a member of ``Cyclopes'', which perfectly enhances answering for the given problem. In the the third example, we can observe that FiD delivers a answer that is only partially correct. Whereas, enhanced by fact relation from KG, \method correctly answers the question, because of the triplet (``UK'', ``applies to jurisdiction'', ``Parliament of UK''). This design is tractable yet effective. Because only entities highly correlated with useful facts will be included. Specifically, passage entities unrelated to question entities are very likely to be marginal and hence removed (i.e., these passage entities are isolated nodes). And only the factual triplets helpful for answering the problems are kept. Relations within passage entities are most likely peripheral and hence neglected in the bipartite graph by definition.

\end{document}